\def\year{2021}\relax
\documentclass[letterpaper]{article} 
\usepackage{aaai21}  
\usepackage{times}  
\usepackage{helvet} 
\usepackage{courier}  
\usepackage[hyphens]{url}  
\usepackage{graphicx} 
\usepackage{natbib}  
\usepackage{caption} 
\def\titleStr{Modeling Deep Learning Based Privacy Attacks on Physical Mail}

\usepackage{hyperref} 

\usepackage{amsmath}
\usepackage{graphicx}
\usepackage{subcaption}
\usepackage{cite}
\usepackage{float}
\usepackage{wrapfig}
\usepackage{longtable,booktabs}
\usepackage{algorithm}
\usepackage[noend]{algpseudocode}
\usepackage{multirow}

\DeclareMathOperator*{\argmin}{arg\,min}

\def\eg{\emph{e.g}.} 
\def\ie{\emph{i.e}.}

\newcommand{\norm}[1]{\left\lVert#1\right\rVert} 

\makeatletter
\newcommand{\thickhline}{%
	\noalign {\ifnum 0=`}\fi \hrule height 1pt
	\futurelet \reserved@a \@xhline
}
\makeatother

\usepackage{fancyhdr}
\fancypagestyle{firstPage}{%
 \fancyhf{}
\chead{\textit{To appear in the 2021 AAAI Conference on Artificial Intelligence (AAAI)}}
\cfoot{1}
}

\title{\titleStr}
\author{
	Bingyao Huang,~~
	Ruyi Lian,~~ 
	Dimitris Samaras,~~
	Haibin Ling\\
}
\affiliations {
	Stony Brook University, NY, USA \\
	\{bihuang, rulian, samaras, hling\}@cs.stonybrook.edu
}

\begin{document}
\maketitle
\thispagestyle{firstPage} 

\begin{abstract}
	Mail privacy protection aims to prevent unauthorized access to hidden content within an envelope since normal paper envelopes are not as safe as we think. In this paper, for the first time, we show that with a well designed deep learning model, the hidden content may be largely recovered without opening the envelope. We start by modeling deep learning-based privacy attacks on physical mail content as learning the mapping from the camera-captured envelope front face image to the hidden content, then we explicitly model the mapping as a combination of perspective transformation, image dehazing and denoising using a deep convolutional neural network, named Neural-STE (See-Through-Envelope). We show experimentally that hidden content details, such as texture and image structure, can be clearly recovered. Finally, our formulation and model allow us to design envelopes that can counter deep learning-based privacy attacks on physical mail.
	\end{abstract}
	
	\section{Introduction}\label{sec:intro}
	With the  recent advances in optical devices and deep learning algorithms, traditional paper envelopes may not be as safe as we think. For example, \cite{redo2016terahertz} show that a \emph{closed} book can be read through using terahertz time-domain spectroscopy; and imaging through scattering media methods \cite{xin2019theory, satat2017object, popoff2010measuring, popoff2010image, kim2015transmission, dremeau2015reference, feng1988correlations, freund1988memory, katz2012looking, katz2014non, bertolotti2012non, judkewitz2015translation,yoon2020deep} can recover the hidden content behind a volume of refractive media, such as ground glass and human tissue. These methods may be extended and applied to attack sealed physical mail.
	
	In this paper, we show that with a carefully designed deep learning model and sampled images captured in a controlled lab environment (see Fig.~\ref{fig:formulation}), the hidden content within the envelope such as the texture and image structure can be recovered without opening it. We start by formulating privacy attacks on physical mail as learning the mapping from the camera-captured envelope image to the hidden content, then we decompose this mapping into a combination of perspective transformation, image dehazing \cite{he2010single, ren2016single} and deblurring \cite{ren2019neural, pan2020physics}. Afterwards, we design three learnable CNN modules to model the three subprocesses. By studying why and how sealed paper mail privacy is subject to deep learning-based attacks, we show that our model can be used to test whether a paper envelope is safe against such attacks and to design safer envelopes in the future.

	For the deep learning-based attacks part, to account for perspective transformations in camera image formation, we use WarpingNet, a module similar to a spatial transformer network (STN) \cite{jaderberg2015spatial}, to warp the camera-captured envelope front face image to the canonical camera frontal view (aligned with the ground truth hidden content). Then, we extend the traditional dehazing formulation \cite{he2010single, ren2016single} to account for the paper envelope's blur operation, transmittance and surface reflectance under the environment light, such that the camera-captured image is a linear combination of the radiance of blurred hidden content, the transmittance of the paper envelope and the reflected light of the surface. Finally, we incorporate such formulation into a deep neural network and explicitly infer these essential intermediate components  using respective CNN modules (\ie, WarpingNet, DehazingNet and RefineNet). This model is trained using sample image pairs of the camera-captured envelope front face (with a hidden printed paper in it) and the ground truth of the printed pattern.	In order to counter hypothetical deep learning-based attacks, we first use our model to test the privacy protection ability of an envelope, and then leverage the envelope properties learned by our attack model to design envelopes that are safe against deep learning-based attacks.

	Our contributions can be summarized as follows:
	\begin{itemize}
		\item The proposed Neural-STE is the first to model deep learning-based privacy attacks on physical mail.
		\item Neural-STE is non-trivially designed as a combination of perspective transformation, image dehazing and deblurring.
		\item Neural-STE can be used to test the privacy-preserving properties of an envelope, and to design envelopes that are safe against deep learning-based attacks.
		\item We propose the first benchmark for modeling privacy attacks on physical mail.  The source code, benchmark dataset and experimental results are publicly available at \url{https://github.com/BingyaoHuang/Neural-STE}.
	\end{itemize}

	\section{Related Work}\label{sec:related_work}
	Our work is most related to \cite{redo2016terahertz} that reads through a closed book using terahertz time-domain spectroscopy. The difference is that we aim to model deep learning-based privacy attacks on physical mail without specialized devices such as a terahertz time-domain system. Moreover, we leverage such an attack model to test envelope safety, and to design safer envelopes. The next related work is \cite{satat2017object} that 
	classifies hidden contents behind the scattering media from the single photon avalanche photodiode (SPAD) captured speckle patterns. It is different from our work since we aim to recover the hidden image rather than classify it. Our work is the first to focus on modeling privacy attacks on physical mail, and to use such a model for mail privacy protection. Another class of related work is imaging through  scattering media that aims to recover hidden contents behind a scattering volume. Finally, our learning-based attack method also relates to image dehazing and deblurring as our deep learning-based attack model incorporates the two formulations in network design.

	\noindent\textbf{Imaging through scattering media~~}
	In this paper, we only review studies on occluding scattering media~\cite{yoon2020deep}, since they relate most to our work. Semi-transparent media such as weather, water, etc., are beyond this paper's scope. Imaging through scattering media can be grouped into traditional optics-based and learning-based methods. 
	Traditional methods use time-resolved measurements \cite{xin2019theory, satat2017object}, transmission matrices \cite{popoff2010measuring, popoff2010image, kim2015transmission, dremeau2015reference} or optical memory effects \cite{feng1988correlations, freund1988memory, katz2012looking, katz2014non, bertolotti2012non, judkewitz2015translation} to reconstruct the hidden scene behind the scattering media. A comprehensive review can be found in \cite{yoon2020deep}.

	Rather than explicitly modeling the light scattering process, learning-based methods \cite{horisaki2016learning, lyu2019learning, li2018imaging, sun2018efficient, guo2020learning, satat2017object, li2018deep} address this issue as an image-to-image translation problem, \ie, translating the sensor-captured speckle patterns to the appearance of the real hidden contents (or projected virtual objects). The first work of this kind \cite{horisaki2016learning} uses pixel-wise support vector regression (SVR) to recover the projected faces behind the scattering media. However, the pixel-wise SVR overfits on faces such that when the testing objects are non-faces the predictions still show strong face patterns. Instead of assuming pixel-wise mapping, deep CNN-based methods \cite{lyu2019learning, li2018imaging, sun2018efficient, guo2020learning, satat2017object, li2018deep} show better accuracy and generalization. 
	
	It is worth noting that our problem is different from imaging through scattering media methods, because sealed pieces of physical mail are different from regular scattering media such as ground glass and human tissue. In addition: 1) imaging through scattering media methods aim to recover accurately the optical properties of a  refractive media volume and the hidden contents behind it, while in our setting we focus on recovering empirical properties of paper envelopes and the hidden contents inside them. Moreover, we show how these empirical properties can be used for mail privacy protection; 2) they usually require specialized optical devices such as such as lasers, projectors, beam splitters, polarizers and SPAD, while we only use a DSLR camera and controllable room lights; and 3) as for data collection they use an additional projector to \emph{project} various sampling patterns onto the back of the scattering media to create sample images, and the model input/output image are geometrically registered, while we manually replace the hidden content within the paper envelopes. In our experiment, we show that directly applying a deep learning-based imaging through scattering media method to attack a piece of physical mail may not work well.

\begin{figure*}[!t]
		\begin{center}
			\includegraphics[width=.8\linewidth]{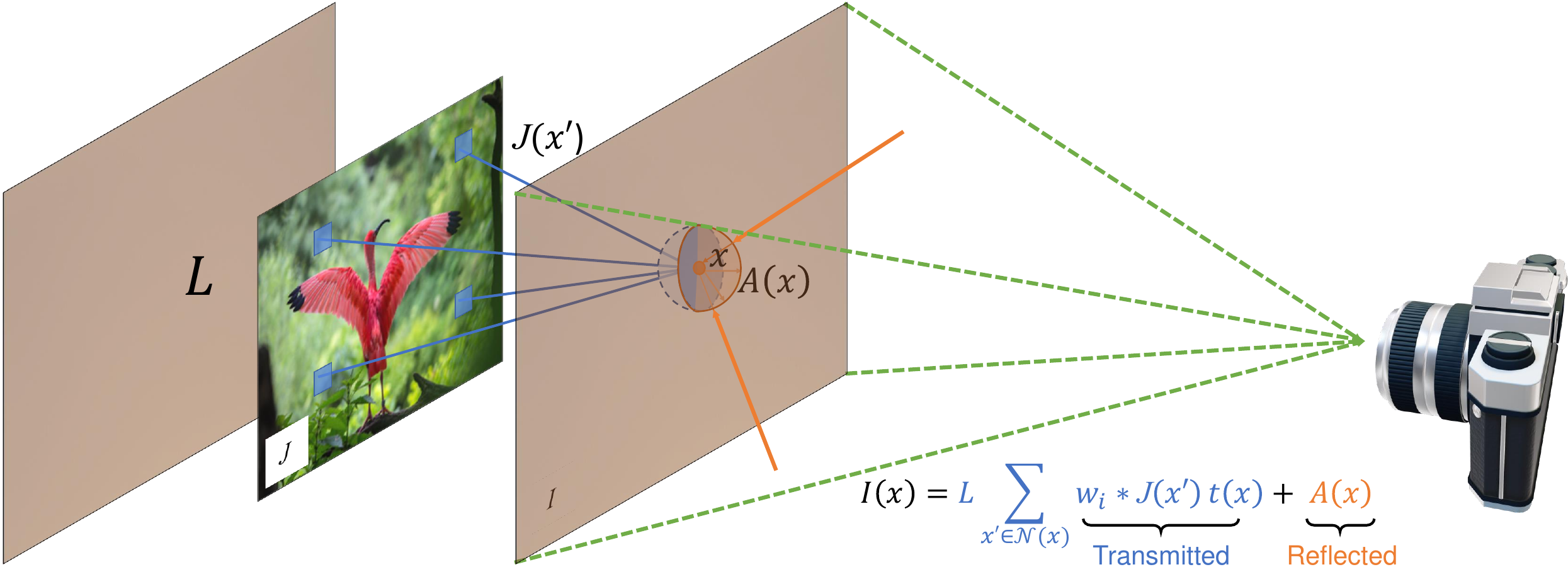}	
			\caption{System setup and empirical image formation model. A color printed paper is put within the envelope (distance exaggerated for illustration). Deep learning-based privacy attacks aim to recover the hidden printed paper $J$ from the camera-captured envelope front surface image  $ I $. Our formulation models $ I $ as a linear combination of the incident environment light $L$, blurred transmitted paper radiance $ J $ and the envelope's front face reflected radiance $ A $. We simplify inter-reflections and subsurface scattering and absorb them in $ A $. 
			}   \label{fig:formulation}	
		\end{center}
	\end{figure*}

	\noindent\textbf{Image dehazing~~}
	Image dehazing \cite{he2010single, ren2016single, pan2020physics} aims to remove haze and reveal the hidden scene radiance. Due to haze \textit{weak} scattering properties, blur can be ignored and the camera-captured image $I$ can be formulated as a linear combination of the haze-free scene radiance $J$, the transmission of the haze $t$ and the atmospheric light $A$:
	\begin{equation}\label{eq:dehaze}
		I(x)=J(x)t(x)+A(1-t(x))
	\end{equation}
	Obviously the equation above cannot be applied to the physical mail attack problem because: (1) Eq.~\ref{eq:dehaze} assumes that each pixel $x$ is independent from other pixel radiance values, which only holds when blur is weak. However, unlike haze, paper envelopes can strongly blur the scene radiance, such that each pixel's radiance is a linear combination (blur/convolution) of its neighboring radiance values (see Fig.~\ref{fig:formulation}). (2) In Eq.~\ref{eq:dehaze}, the transmittance $t$ is assumed uniform across RGB channels and the atmospheric light $A$ is assumed uniform and spatially invariant, but for colored and textured materials with spatially variant microstructures \cite{papas2014physically}, apparently $t$ and $A$ are spatially nonuniform across RGB channels. Such complexity makes explicitly solving for $t$ and $A$ impossible; instead we can infer their empirical versions from sample images using deep learning-based methods.

\noindent\textbf{Image deblurring~~}
	In general, image blurring can be formulated as:
	\begin{equation}\label{eq:deblur}
		I=J\otimes h_x+n
	\end{equation}
	where $I$ is a camera-captured blurred image, $J$ is the latent clean image, $h_x$ is an unknown spatially variant blur kernel, $\otimes$ is the convolution operator and $n$ is the additive noise. This problem can be solved by imposing certain priors on the blurring kernel and noise, such as assuming a known $h_x$ \cite{richardson1972bayesian, lucy1974iterative, gonzales2002digital}, the sparsity of image gradients \cite{chan1998total, fergus2006removing, levin2009understanding} and the dark channel prior \cite{pan2016blind}. Recently, deep image priors \cite{ulyanov2018deep, gandelsman2019double} and GAN-based models \cite{pan2020physics, kupyn2018deblurgan} show clear advantages over previous works. 
	In this paper, inspired by the state-of-the-art learning-based approaches \cite{ulyanov2018deep, gandelsman2019double, pan2020physics, kupyn2018deblurgan}, we show that it is possible to implicitly solve deblurring in our attack model.  
	
	\section{Method}\label{sec:problem}
	\subsection{Problem Formulation}\label{subsec:problem_formulation}
	
	Our setup is shown in Fig.~\ref{fig:formulation}, where a hidden printed paper is placed within an envelope. Following the notation of \cite{he2010single}, denote the camera-captured image as $ I $, the radiance of the hidden printed paper that we aim to recover as $ J $, and the transmittance of the envelope's front face as $ t $. Let $ A $ be the reflected (not transmitted) radiance of the envelope's front face under  normal environment light (\ie, with room lights on).

	Extending Eq.~\ref{eq:dehaze} and Eq.~\ref{eq:deblur} to our problem, the radiance of the camera-captured image $ I $ at pixel $ x $ is given by:
	\begin{equation}
	\begin{split}
		I(x)&=L\sum_{x' \in \mathcal{N}(x)} \omega_iJ(x')t(x)+A(x) \\
		    &=LJ\otimes h_xt(x)+A(x)
	\end{split}
	\label{eq:formulation}
	\end{equation}
	where $L$ is the intensity of the incident environment light on the back face of the hidden content and $ \mathcal{N}(x) $ is a neighborhood of $ x $ and $ \{J(x')|x' \in \mathcal{N}(x)\} $ is a patch of $ J $ centered at $ x $. The transmitted radiance at $ x $ is the weighted sum of all radiance values above the blue hemisphere of $ x $, \ie, $ L\sum_{x' \in \mathcal{N}(x)} \omega_iJ(x') $ (Fig.~\ref{fig:formulation}). This operation can be approximated by convolving $ J $ with a spatially variant convolution kernel $ h_x $, \ie, $ LJ\otimes h_x $, where $ \otimes $ is the convolution operator. Note that the blurring kernel $ h_x $ is also related to the distance between the hidden content and the envelope, \eg, the size of the blurring kernel $ h_x $ increases as the distance increases. We assume that $L$ is a constant scalar and it can be absorbed in $t(x)$. Then, the camera-captured radiance is the sum of the transmitted radiance $ J\otimes h_xt(x)$ and the reflected radiance $ A(x) $ of the envelope. 

	Clearly the problem in Eq.~\ref{eq:formulation} is highly ill-posed, since the unknown $ A, t $ and $ h_x $ are hard to obtain. One intuition is to directly estimate $ J $ from sample image pairs like $ (I, J) $ using an image-to-image translation model \cite{guo2020learning, isola2017image, wang2018pix2pixHD}, however, such general models are not designed for this problem and tend to obtain suboptimal solutions (see Fig.~\ref{fig:qualitative} and Table~\ref{tab:compare}). In this paper, for privacy attacks on physical mail, we decompose this problem as dehazing, deblurring and denoising. First, we estimate the unknowns $ A, t $ using a CNN with some  constraints, then we explicitly compute the blurred radiance by $ J\otimes h_x = (I-A)/t$ as inspired by \cite{he2010single}. Finally, we recover the hidden printed paper radiance $ J $  using an image refinement network to deblur and denoise and to improve color and texture details.
	
	One unaddressed challenge of mail privacy attack is that the above operations assume that the camera-captured image is aligned with the ground truth image as shown in Fig.~\ref{fig:formulation}.
	However, in practice the hidden content $J$ is real printed paper, manually placed within the envelope, thus there is no guarantee that it is aligned with the camera canonical frontal view. Although some CNN-based image-to-image translation models, \eg, Pix2pix \cite{isola2017image} can reconstruct both geometry and colors without explicitly modeling the geometry,  the output may  be subject to a suboptimal solution (color and texture detail loss) when the training samples are limited, as shown in Fig.~\ref{fig:qualitative} columns 2-6  (Fig.~\ref{fig:qualitative}).

	\begin{figure*}[!t]
		\begin{center}
			\includegraphics[width=1\linewidth]{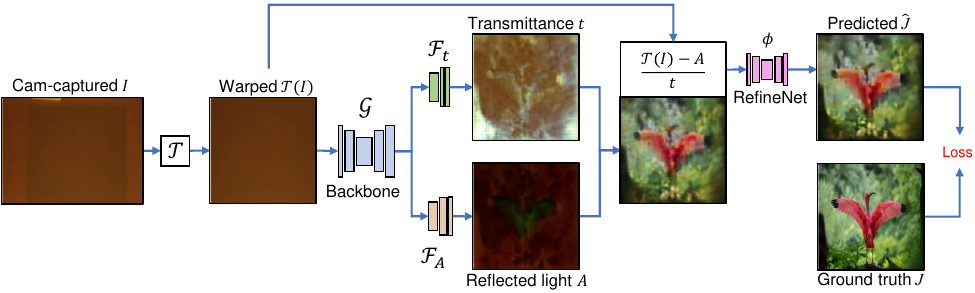}
			\caption{Network architecture of our Neural-STE. It consists of three modules: WarpingNet $\mathcal{T}$, DehazingNet ($\mathcal{G}, \mathcal{F}_A, \mathcal{F}_t$) and RefineNet $\phi$. These modules together with the losses allow us to utilize our image formation model to effectively model privacy attacks on physical mail problem.
			}   \label{fig:net}	
		\end{center}
	\end{figure*}

	To address this issue, we apply a differentiable image warping module to automatically align the input and output images. As shown in our setup Fig.~\ref{fig:formulation}, since for mail, both the envelope and the hidden printed paper are approximately planar, a homography is sufficient to correct the geometric distortions. Thus, we explicitly model the perspective transformation by {an 8-DoF homography}. In summary, our Neural-STE consists of three modules that infer the following intermediate results, respectively:
	\begin{equation}
	\begin{split}
		A &= \mathcal{F}_A(\mathcal{G}(\mathcal{T}(I)))  	\\
		t &= \mathcal{F}_t(\mathcal{G}(\mathcal{T}(I))) 	\\
		\hat{J}  &= \phi((\mathcal{T}(I)-A)/t) \label{eq:J}
	\end{split}	
	\end{equation}
	where $ \mathcal{T} $ is an STN-like \cite{jaderberg2015spatial} homography warping network (Fig.~\ref{fig:net}) that warps the input image to the camera canonical frontal view; $\mathcal{G}$ is a feature extraction function, and $ \mathcal{F}_A $ and $ \mathcal{F}_t $ are two modules that infer the envelope's transmittance $ t $ and its surface reflected light $ A $, respectively. $ \phi $ is an image refinement function which includes image deblurring, denoising and {color and texture refinement}, for conciseness we use image ``refinement"  in the rest of the paper.


	\section{Network Architecture}\label{sec:net}
	Given the formulation in Eq.~\ref{eq:J}, we design our Neural-STE to have three modules (Fig.~\ref{fig:net}): an STN-based \cite{jaderberg2015spatial} \textbf{WarpingNet} that warps the camera-captured image to the camera canonical frontal view (\ie, aligned with the ground truth hidden image $ J $); an encoder-decoder backbone network \textbf{DehazingNet} to infer transmittance $ t $ and surface reflected light $ A $, and compute the coarse dehazed image $ (\mathcal{T}(I)-A)/t $; and a \textbf{RefineNet} to improve texture and color details of the coarse dehazed image.

\begin{figure*}[!t]
		\begin{center}
			\includegraphics[width=1\linewidth]{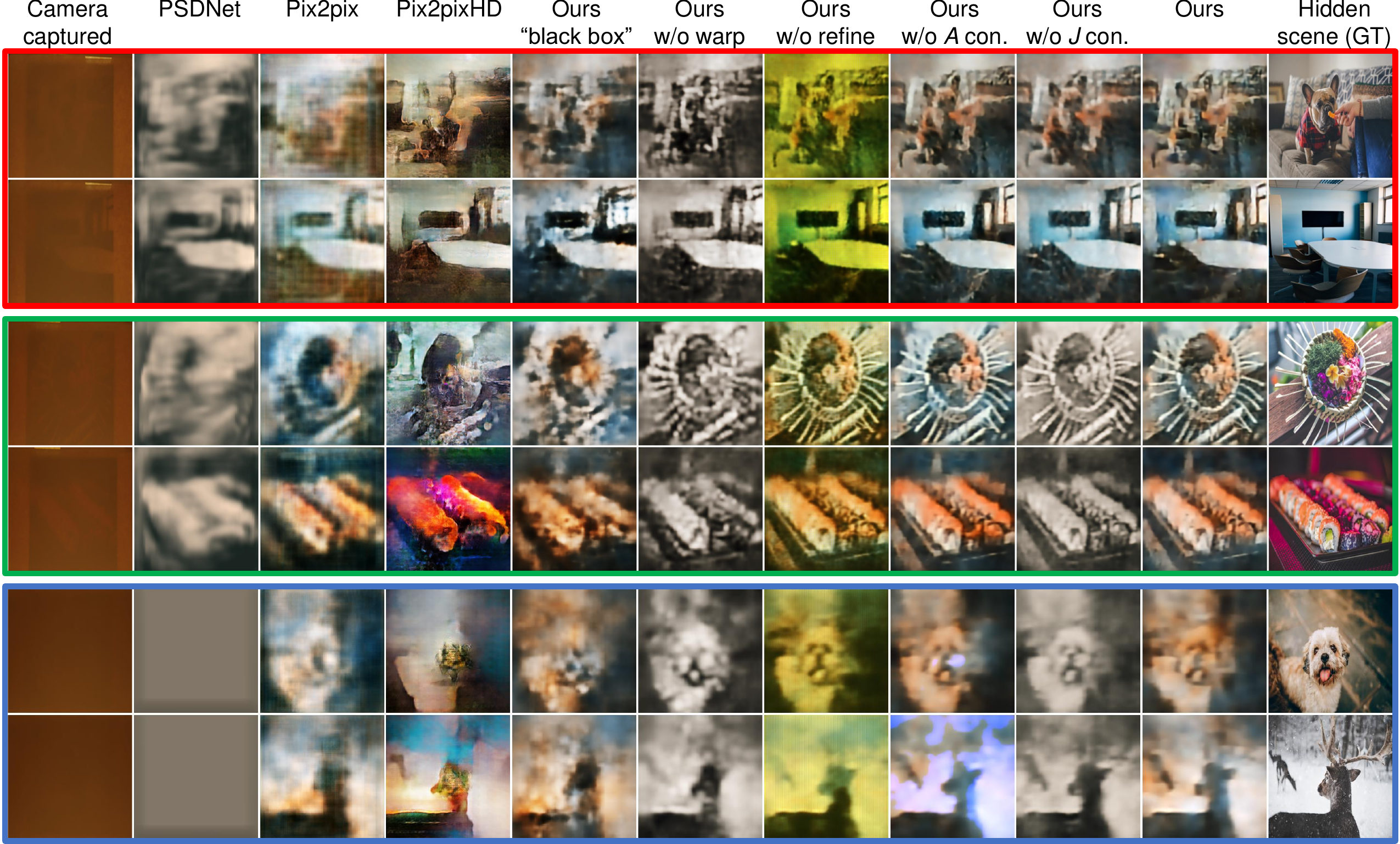}	
			\caption{Qualitative comparison. We show results from three different setups, with the easiest in red and the hardest in blue. We show two examples for each setup and the results of different methods are shown in the 2nd to 9th columns. The 1st column are the camera-captured envelope front face $ I $. The 2nd to 4th columns are PSDNet \cite{guo2020learning}, Pix2pix \cite{isola2017image} and Pix2pixHD \cite{wang2018pix2pixHD}, respectively. The 5th to 9th columns are degraded versions of the proposed Neural-STE, as described in Ablation study. Please see supplementary for larger versions of the images and more results.
			}   \label{fig:qualitative}	
		\end{center}
	\end{figure*}

\noindent\textbf{WarpingNet} ($\mathcal{T}$) is inspired by STN \cite{jaderberg2015spatial} and consists of two convolutional layers, two max pooling layers and two fully connected layers. The module firstly infers a $3\times 3$ homography $H$ from the input image, then warps the input image using this predicted homography by $\mathcal{T}(I) = \texttt{imwarp}(I, H)$\footnote{Taking MATLAB's \texttt{imwarp} as an example.}, such that the warped image $\mathcal{T}(I)$ is roughly aligned with the ground truth hidden image $J$ (Fig.~\ref{fig:net}). Our experimental comparisons show that for our problem, directly learning this geometric transformation is hard if not explicitly modeled as a homography (see \textbf{Ours w/o warp} in Fig.~\ref{fig:qualitative} and Table~\ref{tab:compare}).

\noindent\textbf{DehazingNet} ($\mathcal{G}$, $\mathcal{F}_t, \mathcal{F}_A$) consists of an encoder-decoder backbone network $\mathcal{G}$ and two light weight subnets $\mathcal{F}_t$ and $\mathcal{F}_A$, where the backbone network extracts a 64-dimensional feature map from the input image.
	Then, we design $\mathcal{F}_t$ as a deconvolutional layer and a convolutional layer followed by a sigmoid activation layer, which predicts the envelope's transmittance $ t $. Afterwards, we design $\mathcal{F}_A$ as two convolutional layers followed by a deconvolutional layer and a sigmoid activation layer. Similarly, it predicts the envelope reflected light $A$. Finally the coarse hidden printed paper radiance is computed by $J_{\text{coarse}} = (\mathcal{T}(I)-A)/t $.

\noindent\textbf{RefineNet} ($\phi$). According to our formulation in Eq.~\ref{eq:formulation}, recovering the hidden content within a physical mail is more than just dehazing; we need to account for other deformations such as blurring and noise. Note that instead of explicitly estimating the blur kernel, inspired by the success of recently proposed learning-based approaches \cite{ulyanov2018deep, gandelsman2019double, pan2020physics, kupyn2018deblurgan}, we design \textbf{RefineNet} as a CNN with a skip connection \cite{he2016deep}, such that the color and texture details can be learned as a residual image. In Fig.~\ref{fig:intermediate}, comparing the refined image $\hat{J} = \phi(J_{\text{coarse}})$ with the coarse image $J_{\text{coarse}}$, we see clearly improved details.
	
\begin{table*}[!t]
	\begin{center}
	\begin{tabular}{lccc c lccc}
		\toprule[0.5mm]
		\textbf{Model}                     & \textbf{PSNR}$\uparrow$ & \textbf{RMSE}$\downarrow$ & \textbf{SSIM}$\uparrow$ &  & \textbf{Model (degraded)} & \textbf{PSNR}$\uparrow$ & \textbf{RMSE}$\downarrow$ & \textbf{SSIM}$\uparrow$ \\ \midrule[0.3mm]
		Cam-captured                       &         ~8.2767         &          0.6682           &         0.2695          &  & Ours ``black box''        &         14.1106         &          0.3442           &         0.3914          \\
		PSDNet \cite{guo2020learning}      &         12.8952         &          0.3981           &         0.3717          &  & Ours w/o refine           &         11.4025         &          0.4696           &         0.3034          \\
		Pix2pix \cite{isola2017image}      &         12.2620         &          0.4238           &         0.3409          &  & Ours w/o warp             &         14.3415         &          0.3345           &         0.4125          \\
		Pix2pixHD \cite{wang2018pix2pixHD} &         12.0964         &          0.4303           &         0.3193          &  & Ours w/o A con.           &         14.7582         &          0.3239           &         0.4421          \\
		Neural-STE (ours)                         &    \textbf{15.0275}     &      \textbf{0.3127}      &     0.4449     &  & Ours w/o J con.           &         14.9082         &          0.3151           &         \textbf{0.4460}          \\ \bottomrule[0.5mm]
	\end{tabular}
	\caption{Quantitative comparison. Results are averaged over three setups, each containing 50 testing images. ``Cam-captured" is the similarity between the camera-captured envelope front face and the ground truth. See supplementary for separate measurements for each setup. 
	}\label{tab:compare}
	\end{center}
\end{table*}

\noindent\textbf{Additional constraints.} 
	With the three parameterized modules, training image pairs and a proper loss function $\mathcal{L}$, the network parameters can be learned by:
	\begin{equation}
	\begin{split}
		& \{\mathcal{T}, \mathcal{G},\mathcal{F}_t,\mathcal{F}_A, \phi\}_\theta = \\
		& ~~~~~~ \argmin\mathcal{L}(\hat{J} = \phi((\mathcal{T}(I)-A)/t), \ J) \label{eq:train}
	\end{split}
	\end{equation}
	However, directly training this network may lead to a suboptimal solution and the output images may contain strange color or lack color {(the 8th-9th columns of Fig.~\ref{fig:qualitative})}. In our setup, we have observed that the surface reflected light $A$ dominates the warped camera-capture image $\mathcal{T}(I)$ (\ie, contributes more light than the transmitted light) due to the ambient light, thus we assume that $A$ should look like $\mathcal{T}(I)$. Then we impose this constraint as a pixel-wise $L_2$ loss {$\norm{\mathcal{T}(I) - A}_2^2$}. Moreover, the computed coarse result {$J_{\text{coarse}}=(\mathcal{T}(I)-A)/t$} should look like the ground truth $J$, except for some differences in color and texture details, for which we introduce a pixel-wise $L_2$ loss {$\norm{J - J_{\text{coarse}}}_2^2$}.

	Another constraint is that the paper envelopes are not fully opaque ($t\not\approx0$), otherwise not only the problem is meaningless, but also Eq.~\ref{eq:J} may be divided by zero. Thus, we clip the transmittance $t$ to $[0.01, 1]$.

\noindent\textbf{Loss function.}
	As shown in Eq.~\ref{eq:total_loss}, our loss function consists of three terms, an image reconstruction loss $\mathcal{L}_{\text{recon}}$ (Eq.~\ref{eq:recon}), \ie, the pixel-wise $L_{1}+\text{SSIM}$ loss \cite{zhao2017loss} between the predicted hidden printed paper content image $\hat{J}$ and the ground truth $J$; and the two constraint-based losses above. Then our deep learning-based attack model is trained using Eq.~\ref{eq:train} with the loss below.
	\begin{align}
		&\mathcal{L}  = \mathcal{L}_{\text{recon}}(J, \hat{J}) + \norm{J\! -\! J_{\text{coarse}}}_2^2   + 0.1\norm{A\!-\!\mathcal{T}(I)}_2^2 \label{eq:total_loss} \\
		&\mathcal{L}_{\text{recon}}(J, \hat{J}) = |J-\hat{J}| + 1-\text{SSIM}(J, \hat{J}) \label{eq:recon}
	\end{align}

\noindent\textbf{System configuration and implementation.}
	The proposed setup consists of a Canon 6D camera with the resolution set to 320$\times$240. We color print 500 colorful textured images at US letter size as ground truth hidden contents. Unlike imaging through scattering media, for each capture, we manually replace the printed paper within the envelopes and this operation requires touching the envelopes, thus the shape and pose of the hidden printed paper and the envelopes are inevitably changed each time, making the  hidden content recovery more difficult. The distance between the camera and the envelope is around one meter. The only light sources are various room lights. The collected data is available as our Neural-STE dataset. We implement Neural-STE using PyTorch \cite{paszke2017automatic} and Kornia \cite{eriba2019kornia}, and optimize it using the Adam optimizer \cite{kinga2015method}. The initial learning rate and penalty factor are set to $ 10^{-3} $ and $ 5*10^{-4} $, respectively. Then, we train the model for 4,000 iterations on three Nvidia GeForce 1080Ti GPUs with a batch size of 16, taking about 18 minutes to train.

	\section{Experimental Evaluations}\label{sec:experiments}
	
	\subsection{Privacy Attacks on Physical Mail}
	In this section, we  quantitatively and qualitatively evaluate and compare the proposed Neural-STE with PSDNet \cite{guo2020learning}, a learning-based image through scattering media method, Pix2pix \cite{isola2017image}, a general GAN-based image-to-image translation model, Pix2pixHD \cite{wang2018pix2pixHD}, an improved version of Pix2pix, and degraded versions of the proposed method.

\noindent\textbf{Evaluation benchmark.}
	We prepared two different sets of envelopes and three different setups, which were configured to cover three levels of difficulty. As shown in Fig.~\ref{fig:qualitative}, the red box shows a thin kraft envelope imaged under bright room light. The green box shows a thick kraft envelope imaged under bright room light; and the blue box shows a thick kraft envelope under normal room light. 
	For each setup, we split the captured 500 image pairs into 450 training samples and 50 testing samples. Then, the hidden contents $\hat{J}$ recovered by different methods are compared with the ground truth $J$ using PSNR, RMSE and SSIM \cite{wang2004image}.

	Since our method is the first to model privacy attacks on physical mail, there is no previous work to compare with. Instead, we compare with PSDNet \cite{guo2020learning}, an image through scattering media method. The mapping from the camera-captured image to the hidden content radiance is directly learned without an explicit image formation model like us. While the original PSDNet is designed to only work for grayscale images, we extend it to RGB by increasing the input and output channels.  As shown in Fig.~\ref{fig:qualitative}, PSDNet is unable to recover the hidden contents on the hardest setup (the blue box).  Please see the supplementary material for comparisons on grayscale images.
	
	We then compare with a GAN-based general image-to-image translation network Pix2pix \cite{isola2017image} and its improved version Pix2pixHD \cite{wang2018pix2pixHD}. We train them for 23,000 iterations with a batch size of one. Note that Pix2pix and Pix2pixHD have more parameters than our model, yet they cannot generate satisfactory results as shown in Fig.~\ref{fig:qualitative} and Table~\ref{tab:compare}, because they are designed for general image-to-image translation with a relatively large training dataset (both number and diversity), and they may not work well for the privacy attacks on physical mail setting when the  training data is limited.

\noindent\textbf{Ablation study.}
	To show the effectiveness of our formulation and network architecture, we compare the proposed Neural-STE with its degraded versions, each with a certain module or constraint removed. For example, \textbf{Ours ``black box''} is a naive UNet-like \cite{ronneberger2015u} model without WarpingNet $\mathcal{T}$, DehazingNet\footnote{Note that $\mathcal{F}_t$ and $ \mathcal{F}_A$ are concurrently used to dehaze the image, thus we disable them together for this degraded version.} $\mathcal{F}_t, \mathcal{F}_A$ or RefineNet $\phi$ and the camera-captured image is resized to 256$\times$256 then input to the backbone network $\mathcal{G}$. Moreover, rather than explicitly computing the dehazed and refined image using Eq.~\ref{eq:J}, the output image is predicted by a three-channel convolutional layer concatenated to the backbone network. 
	\textbf{Ours w/o warp} is Neural-STE without WarpingNet $\mathcal{T}$, and the camera-captured image is also resized to 256$\times$256 before fed to the backbone network $\mathcal{G}$. 
	\textbf{Ours w/o refine} is the same as Neural-STE but with RefineNet $ \phi $ removed. 
	\textbf{Ours w/o A con.} and \textbf{Ours w/o J con.} are Neural-STE without the constraints on $ A $ (\ie, $ 0.1\norm{A-\mathcal{T}(I)}_2^2 $ in Eq.~\ref{eq:total_loss}), and $ J $ (\ie, $ \norm{J - J_{\text{coarse}}}_2^2 $ in Eq.~\ref{eq:total_loss}), respectively. 
	
	The experimental comparisons in Table~\ref{tab:compare} and Fig.~\ref{fig:qualitative} clearly show that the proposed Neural-STE outperforms degraded versions that only model part of our image formation process. For example, comparing the 2nd to the 6th columns with the 7th to the 9th columns in Fig.~\ref{fig:qualitative}, it is clear that explicitly modeling the perspective transformation is important for this problem, because when the inferred image is aligned to the ground truth (especially in the early training stages), the model may focus on refining only the color and texture details, which reduces the probability of falling into local minima early on and improves convergence. Note that \textbf{Ours w/o refine} has yellowish output because it cannot fully remove the envelope surface color. \textbf{Ours w/o A con.} and \textbf{Ours w/o J con.} work well for the easiest envelope, but \textbf{Ours w/o J con.} fails to recover hidden scene colors from the other two harder envelopes, and \textbf{Ours w/o A con.} generates unwanted bluish pattern for the hardest envelope.

	\begin{figure}[!t]
		\begin{center}
			\includegraphics[width=1\linewidth]{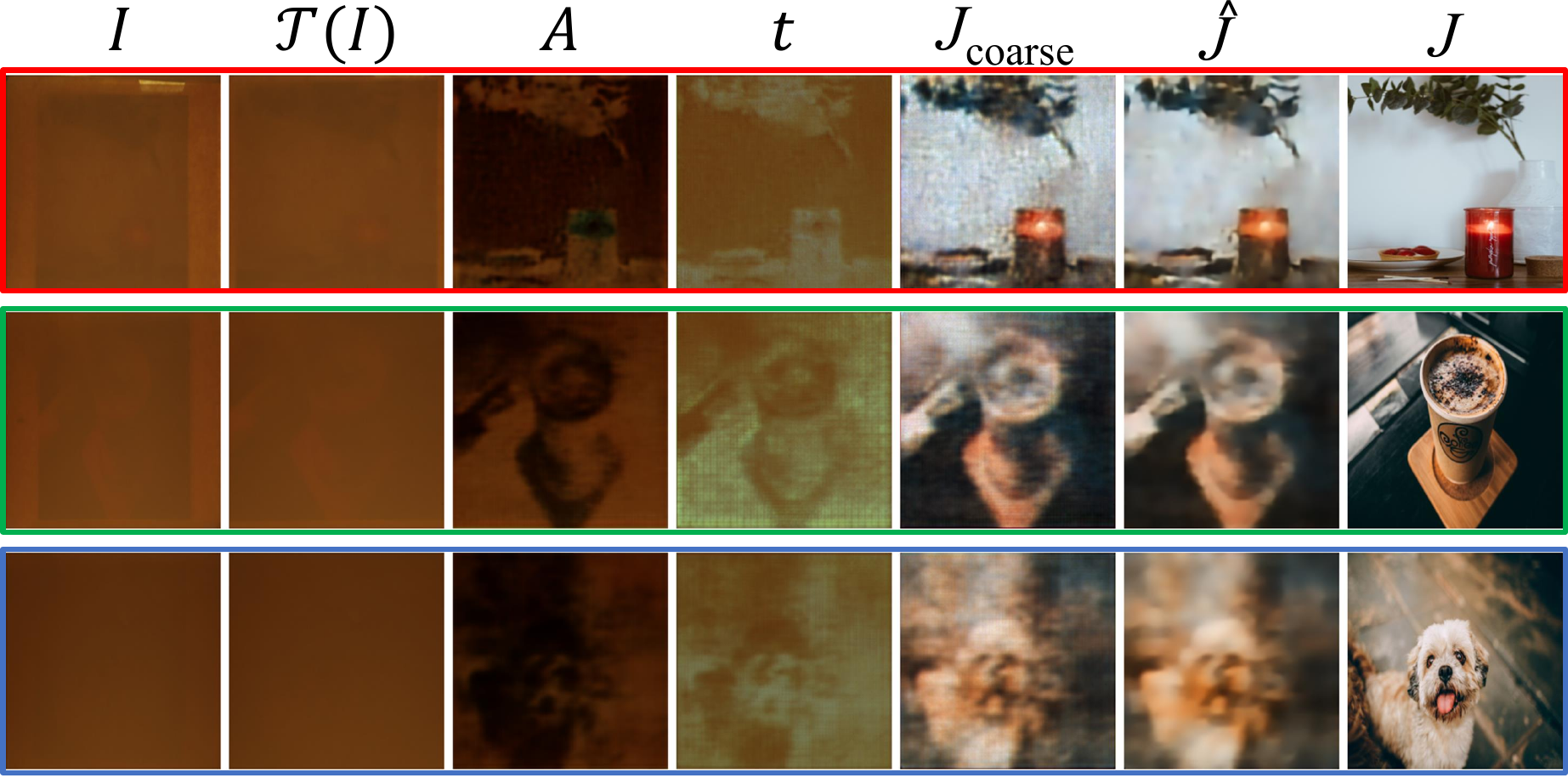}	
			\caption{Intermediate results of Neural-STE. See supplementary for more images of the three setups.
			}   \label{fig:intermediate}
		\end{center}
	\end{figure}
	
	We then show the intermediate results of Neural-STE in Fig.~\ref{fig:intermediate}. The red and the blue boxes are the easiest and the hardest setups, respectively. The columns are, from left to right, the camera-captured image $I$, the WarpingNet warped image $ \mathcal{T}(I)$, the predicted reflected light $A$, the predicted transmittance $t$, the coarse dehazed hidden content radiance $J_{\text{coarse}} = (\mathcal{T}(I)-A)/t $, the final refined results (\ie, $\hat{J} = \phi(J_{\text{coarse}}) $), and the ground truth hidden content $ J $. 	These empirical properties can then be used to design safer envelopes that counters deep learning-based attacks below.
	
	\subsection{Countering Privacy Attacks on Physical Mail}\label{sec:privacy}
	In this section, we show how to leverage our empirical image formation model and our Neural-STE to design envelopes that can defend mail privacy against deep learning-based attacks. We conducted both simulated and real experiments.
	
	\noindent\textbf{Simulated experiments.}
	We simulated camera-captured images using our empirical image formation model in Eq.~\ref{eq:formulation} and intermediate results of our attack method. The environment light $L$, the perspective transformation $H$, the blur kernel size $|h_x|$, the transmittance coefficient\footnote{Note that $t$ and $A$ are maps, here we use coefficients (scalars) to control their strengths, \eg, $k_t*t$ and $k_A*A$.} $k_t$, the surface reflected light coefficient $k_A$ were varied in simulation. In addition, we added Gaussian white noise to the blurred image and Poisson noise to the final camera-captured image. 

\begin{figure}[!t]
		\begin{center}
	   \includegraphics[width=1\linewidth]{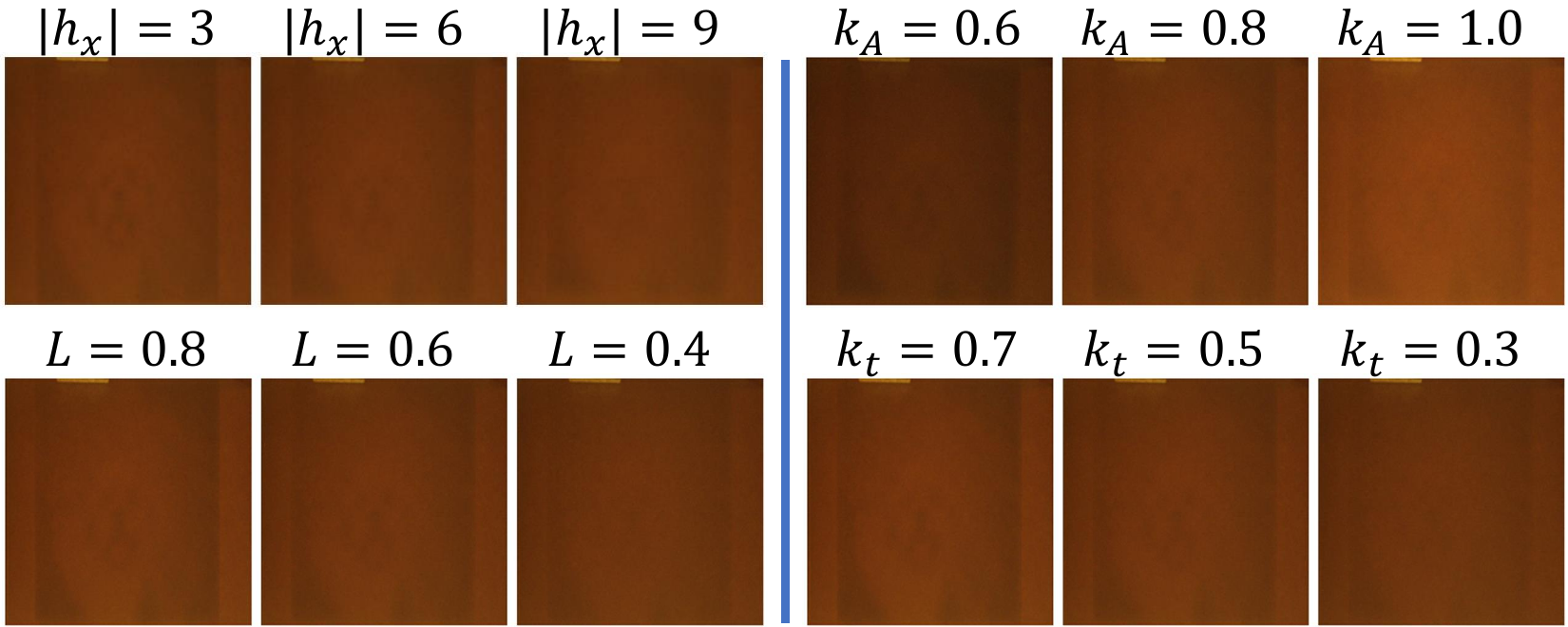} 
	   \caption{Visualization of the simulated camera-captured image when we tune the controllable parameters,  \ie, the size of the blur kernel $h_x$, the surface reflected light $A$, the environment light $L$ and the envelope's transmittance $t$. Here we show an easy setup so that the hidden content is recognizable: as we tune each optical parameter from left to right, the hidden content becomes harder to recognize. 
	   } \label{fig:simulation_param}
	   \end{center}
	\end{figure}

	Then, we simulated nine synthetic setups, \ie, for each of the controllable parameters, we generated three variations. In Fig.~\ref{fig:simulation_param}, we show how each controllable parameter affects appearances of the final simulated camera-captured images. Afterwards, we applied our Neural-STE to the simulated dataset; an example result is shown in Fig.~\ref{fig:simulated_counter_attack}. See supplementary for quantitative comparisons. We first applied our Neural-STE to examine  envelope security. For example, if the envelope can be successfully attacked (as shown in Fig.~\ref{fig:simulated_counter_attack}, \textbf{Unsafe envelope}), we redesign the envelope using our empirical image formation model, until our Neural-STE fails. In our experiment, we find that, safer envelopes can be manufactured by using a material that has a more reflective surface (\ie, larger $A$); by increasing the envelope thickness or using a material that absorbs more light (\ie, smaller transmittance $t$); and by increasing the distance between the hidden contents and the envelope (\ie, a larger blur kernel $h_x$). For example, in Fig.~\ref{fig:simulated_counter_attack}, \textbf{Safe envelope} has the following properties: $|h_x|=17, k_A=1.0, k_t=0.1$.

	\begin{figure}[!h]
		\begin{center}
	   \includegraphics[width=1\linewidth]{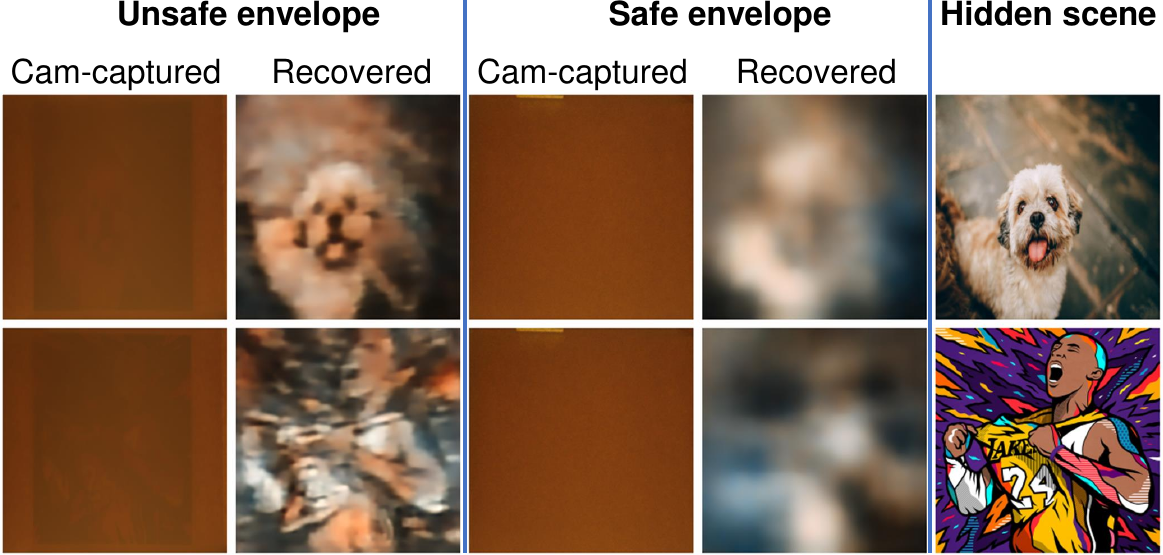} 
	   \caption{Countering hypothetical deep learning-based privacy attacks on simulated physical mail.} \label{fig:simulated_counter_attack}
	   \end{center}
	\end{figure}

	\begin{figure}[!t]
		\begin{center}
			\includegraphics[width=1\linewidth]{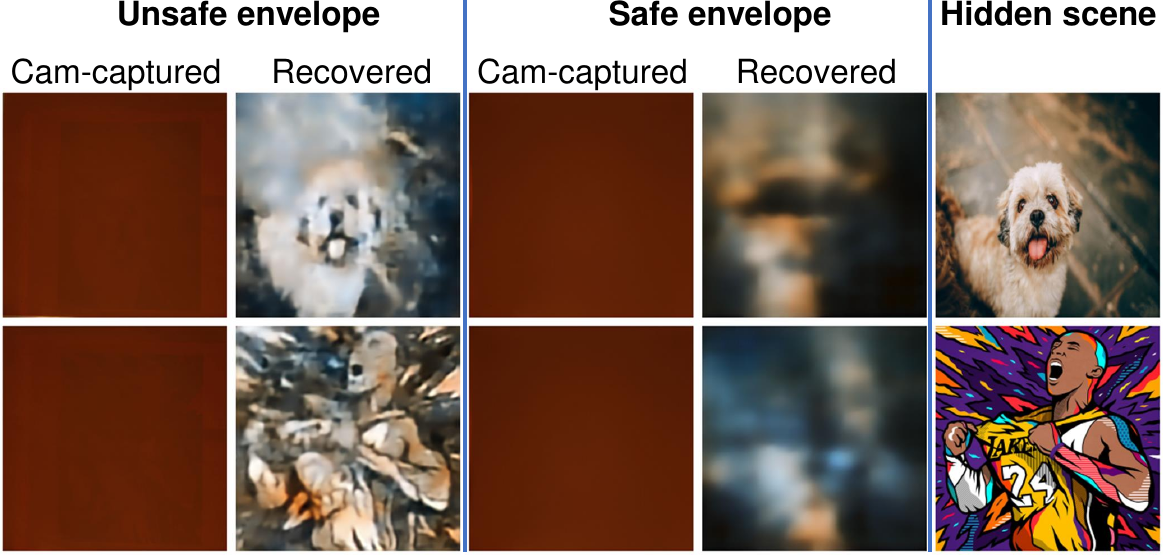}	
			\caption{ Countering hypothetical deep learning-based privacy attacks on real physical mail.
			}   \label{fig:real_counter_attack}
		\end{center}
	\end{figure}

	\noindent\textbf{Real experiments.} 
	We prepared a new real envelope and attacked it with the proposed Neural-STE, as shown in Fig.~\ref{fig:real_counter_attack}, \textbf{Unsafe envelope}. Then, according to the findings in simulation results, we placed an additional paper layer in the envelope to reduce the transmittance $ t $ and to increase the blur kernel $ h_x $, and we named it \textbf{Safe envelope}. Afterwards, we attacked it using Neural-STE, and clearly our method failed to reveal the hidden contents within it.
	
	\section{Conclusion and Limitations}\label{sec:conclusion}
	In this paper, we proposed the first deep learning-based method, named Neural-STE for privacy attacks on physical mail. Our method explicitly decomposes an empirical image formation model into a combination of perspective transformation, blurring, transmission and surface reflected light, and then we non-trivially designed respective CNN modules to learn these intermediate results. We proposed the first privacy attacks on physical mail benchmark  and we expect it to facilitate future work in this direction. Our experimental results on this benchmark clearly show that normal envelopes are not as safe as we think. Finally, we leverage the empirical image formation model and Neural-STE to design envelopes that can counter such attacks.
	
	\noindent\textbf{Limitations.} We have only tested our Neural-STE on kraft envelopes and it may not work well on other materials with much stronger scattering properties. Moreover, we have not tested the method for privacy attacks on folded text, which is much more challenging.
	Extending this method to (counter) such attacks is definitely an interesting direction to explore.
	
	\noindent\textbf{Acknowledgements.} This work was partially supported by the Partner University Fund, the SUNY2020 ITSC, a gift from Adobe, the Yahoo Faculty Research and Engagement Program Award, and the US NSF Grants 2006665.

	\bibliography{ref}
\end{document}